\documentclass[letterpaper, 10 pt, conference]{ieeeconf}

\usepackage{amsmath}
\usepackage{amssymb}
\usepackage{mathtools}

\usepackage{tabularx}
\usepackage{graphicx}
\usepackage{bm}
\usepackage{color}
\usepackage{float}
\usepackage{units}
\usepackage{url}
\usepackage{hyperref}
\usepackage{balance}
\usepackage{amsmath}
\usepackage{makecell}

\makeatletter 
\def\endfigure{\end@float} 
\def\endtable{\end@float}
\makeatother

\usepackage{subcaption}
\usepackage[]{graphicx}
\usepackage{caption}

\renewcommand{\unit}[1]{{\rm #1} }

\IEEEoverridecommandlockouts

\begin{document} 

\title{\Large \bf 			
Continuous Dynamic Bipedal Jumping via Real-time Variable-model Optimization
}

\author{Junheng Li, Omar Kolt, and Quan Nguyen\thanks{Junheng Li, Omar Kolt, and Quan Nguyen are with the Department of Aerospace and Mechanical Engineering, University of Southern California, Los Angeles, CA 90089.\endgraf
Email: {\tt\small $\{$junhengl, kolt, quann$\}$@usc.edu}. \endgraf
This work is supported by USC Departmental Startup Fund. }}

\maketitle


\begin{abstract}
Dynamic and continuous jumping remains an open yet challenging problem in bipedal robot control. 
Real-time planning with full body dynamics over the entire jumping trajectory presents unsolved challenges in computation burden. 
In this paper, we propose a novel variable-model optimization approach, a unified framework of variable-model trajectory optimization (TO) and variable-frequency Model Predictive Control (MPC), to effectively realize continuous and robust jumping planning and control on HECTOR bipedal robot in real-time. The proposed TO fuses variable-fidelity dynamics modeling of bipedal jumping motion in different jumping phases to balance trajectory accuracy and real-time computation efficiency. 
In addition, conventional fixed-frequency control approaches suffer from unsynchronized sampling frequencies, leading to mismatched modeling resolutions. We address this by aligning the MPC sampling frequency with the variable-model TO trajectory resolutions across different phases.
In hardware experiments, we have demonstrated robust and dynamic jumps covering a distance of up to 40 cm (57$\%$ of robot height). To verify the repeatability of this experiment, we run 53 jumping experiments and achieve 90$\%$ success rate.  In continuous jumps, we demonstrate continuous bipedal jumping with terrain height perturbations (up to 5 cm) and discontinuities (up to 20 cm gap). 

\end{abstract}


\section{Introduction}
\label{sec:Introduction}

Bipedal robots have demonstrated a large step of advancement in dynamic, adaptive, and robust locomotion in recent years \cite{nguyen2018dynamic, kim2020dynamic, li2021reinforcement, gibson2022terrain, gu2024robust}. Traditional approaches in bipedal locomotion control rely on periodic walking gait to ensure stability and always in-contact \cite{vukobratovic2001zero, vaughan2003theories}. 
Boston Dynamics and Unitree Robotics have showcased impressive parkour and back-flipping motions on full-size humanoids \cite{BostonDynamics_2021, Unitree_2024}, demonstrating the capability of bipedal/humanoid robots in locomotion techniques with extended flight phases and beyond walking gaits.
With the motivation of allowing bipedal robots to efficiently plan and execute dynamic jumping motions in real-time with adaptivity and agility like humans, in this work, we propose a robust and versatile bipedal jumping motion planning and control framework.


Model-based control approaches to solving quadruped jumping motion generation have a strong presence in recent literature and usually involve solving offline trajectory optimization (TO) problems \cite{nguyen2019optimized,chignoli2022rapid,ding2020kinodynamic,bellegarda2024quadruped}. 
Due to the inherently unstable and under-actuated nature of bipedal robots, bipedal jumping consists of more challenges in model-based control system design, such as in addressing (1) leveraging whole-body motion for takeoff, (2) effective pose tracking during the flight phase for optimal landing configuration, (3) impact mitigation and balancing upon landing, (4) real-time planning, and (5) transferability to hardware.

\begin{figure}[!t]
\vspace{0.2cm}
     \centering
     \begin{subfigure}[b]{0.48\textwidth}
        \includegraphics[clip, trim=0cm 10.2cm 0cm 0cm, width=1\columnwidth]{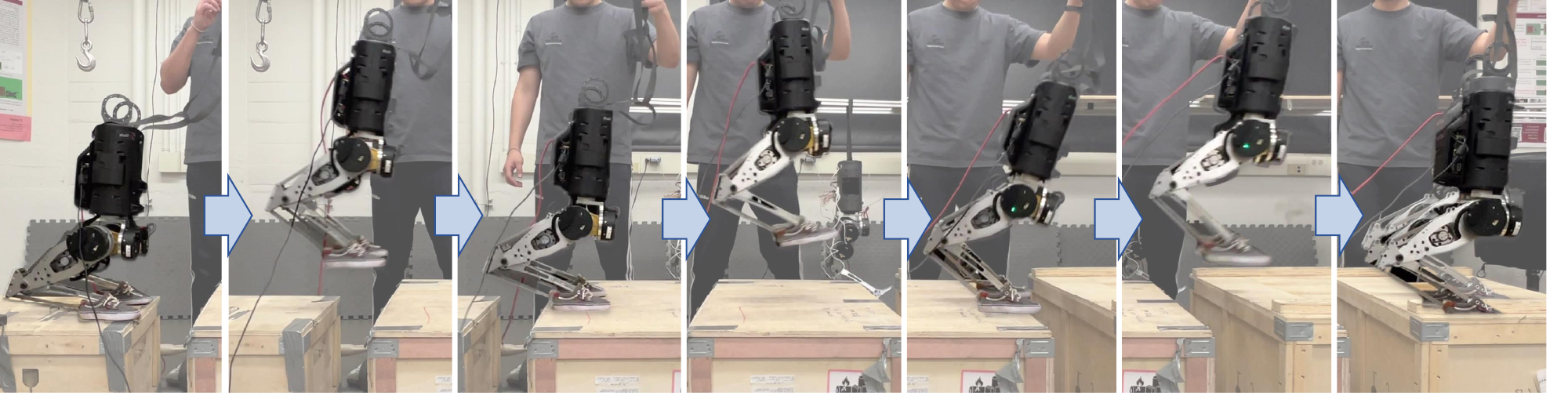}
	\caption{Continuous jumping over discrete terrain.}
        \vspace{0.2cm}
	\label{fig:continuous}
     \end{subfigure}
     \begin{subfigure}[b]{0.46\textwidth}
         \centering
	   \includegraphics[clip, trim=0cm 9.5cm 0cm 0cm, width=1\columnwidth]{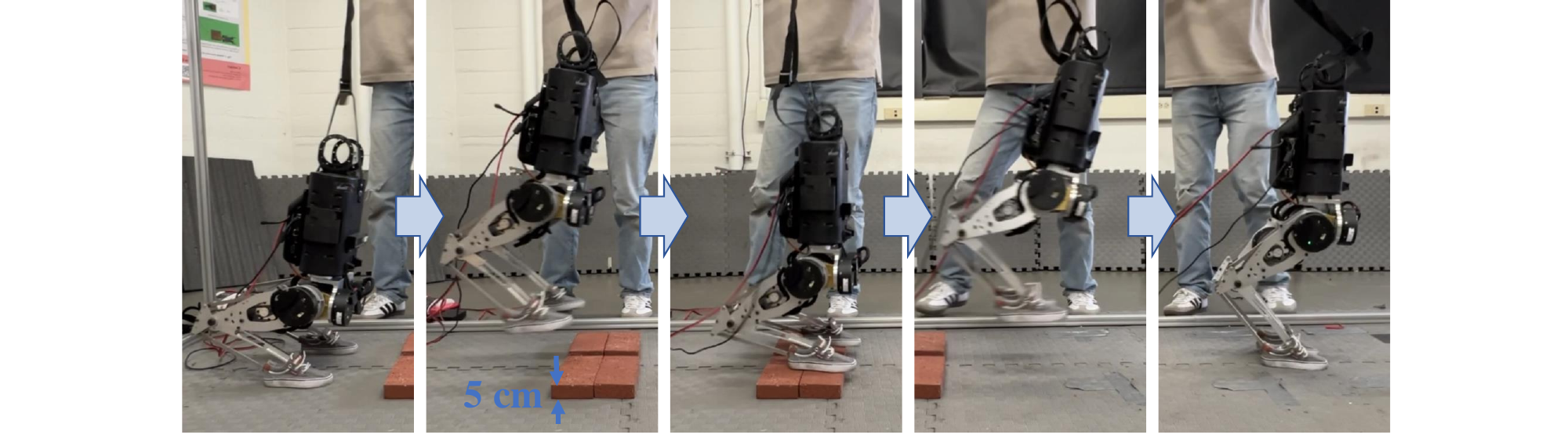}
          \caption{Continuous jumping with terrain perturbation.}
          \vspace{0.1cm}
          \label{fig:jumpdown}
     \end{subfigure}
      \begin{subfigure}[b]{0.46\textwidth}
         \centering
	   \includegraphics[clip, trim=0cm 0cm 0cm 9.5cm, width=1\columnwidth]{Fig/title2.pdf}
          \caption{Jumping down from 25 cm block.}
        \vspace{-0.0cm}
        \label{fig:terrain}
     \end{subfigure}
     \caption{{\bfseries{Dynamic and Continous Bipedal Jumping on HECTOR.}} Full experiment video: \url{https://youtu.be/TDzxay3PuEM}}
        \label{fig:title}
\end{figure}

\begin{figure*}[!t]
\vspace{0.2cm}
		\center
		\includegraphics[clip, trim=0.5cm 0.9cm 1cm 0.8cm, width=2\columnwidth]{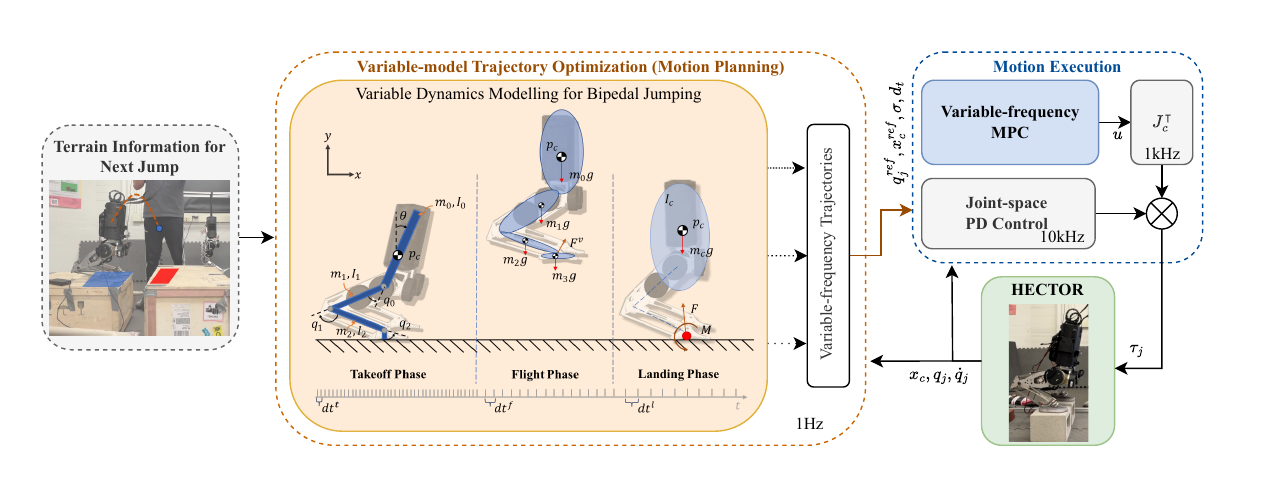}
		\caption{{\bfseries System Architecture}}
		\label{fig:controlArchi}
		\vspace{-0.2cm}
\end{figure*}


Related works have made attempts to address the above challenges partially.
For instance, many related work leverage kino-dynamics modeling in bipedal jumping planning\cite{chignoli2021humanoid,zhang2023design,he2024cdm}, which considers whole-body link effects with reduced computation burden. However, it still presents computation efficiency challenges for real-time usage. Additionally,
Xiong et al. leverages a single spring-mass model to plan bipedal hopping and landing trajectories via direct collocation \cite{xiong2018bipedal}. Hierarchical TO is proposed in \cite{nguyen2022contact} to uses TO simple model first to quickly generate contact timings to reduce additional computation cost in subsequent TO with full-order dynamics. 
A common theme among these related works is utilizing a single modeling method in TO as an offline motion planner. As for the modeling choice, on one hand, reduced-order modeling such as kino-dynamics assumes the robot as a single-rigid-body dynamics (SRBD) model or centroidal dynamics (CD) model \cite{orin2013centroidal} with kinematic constraints \cite{herzog2016structured}. On the other hand, whole-body dynamics TO utilizes a full-order dynamics model of the robot and contact dynamics for very accurate trajectory planning but may take significantly longer time to generate \cite{posa2014direct}.
 
In this work, we attempt to address the above problems altogether by finding a well-balanced middle ground and proposing a Variable-model TO framework that takes into consideration the model fidelity requirement in TO during different jumping phases to balance the resolution of the trajectory and computation effort. The proposed framework can offer real-time, accurate, and hardware-realizable trajectories.

Allowing variability of model resolutions in optimization problems has become a popular strategy in model-based control. 
Li et al. propose a Model Hierarchical Predictive Control that prioritizes high-resolution models in near MPC horizons and uses reduced-order models in far prediction horizons \cite{li2021model,li2024cafe}. Csomay-Shanklin et al. use a finer sampling rate during contact and a relaxed sampling rate during the flight phase on monoped robots. \cite{csomay2023nonlinear}. 

In addition, conventional model-based tracking control strategies use fixed-frequency control that is unsynchronized with planned trajectories (\textit{e.g.}, \cite{nguyen2022contact,chignoli2021humanoid}), which does not fully leverage the model resolution for effective tracking performance. 
In this work, we validate the significance of matching frequency in planned trajectory and control through an SRBD-based variable-frequency MPC, inspired by prior work \cite{li2023dynamic}.

The main contributions of the paper are as follows:
\begin{itemize}

    \item We propose a real-time bipedal jumping planner using a variable-model trajectory optimization (TO) framework with three dynamic models and sampling rates, each tailored to the takeoff, flight, and landing phases.

    \item We propose a variable-frequency scheme to allow matching frequency between planned trajectory and motion execution (\textit{i.e.,} MPC) to fully exploit the synergy of trajectory resolution among the control hierarchy.

    \item We introduce a landing recovery trajectory in bipedal jumping to complement the variable-model TO and address the rebalancing and recovery challenge after the landing. 
    
    \item In hardware validations, our proposed framework demonstrated robust and repeatable jumping planning and execution. We also showcase the first instance of dynamic continuous bipedal jumping on hardware over discrete terrains with optimization-based control strategies.

\end{itemize}

The rest of the paper is organized as follows. Section. \ref{sec:overview} presents the overview of the proposed control system architecture. Section.\ref{sec:approach} introduces the details of the bipedal robot models used in variable-frequency TO, its nonlinear programming (NLP) problem definition, and the variable-frequency MPC. Section. \ref{sec:Results} presents the numerical comparative analysis and experimental validations.

\section{System Overview}
\label{sec:overview}

In this section, we introduce the system architecture of the proposed control architecture, shown in Fig. \ref{fig:controlArchi}. The proposed Variable-model Optimization framework consists of two major parts, the variable-model TO and vairable-frequency MPC. 

Conventional whole-body dynamics TO with contact is computationally intensive, particularly for high-DoF bipedal robots. Even reduced-order methods like kino-dynamic TO face challenges in achieving real-time planning. The variable-model TO addresses this by using multiple dynamics models and variable sampling rates, efficiently generating real-time, hardware-realizable jumping trajectories.

The framework first generate target landing locations based on the \textit{one-step} terrain data or user input. The proposed variable-model TO serves as a motion planner and generates variable-frequency jumping trajectories.
With $N$ time-steps, these trajectories are in terms of 2D reference trunk Center of Mass (CoM) states, $\mathbf{x}_c^{\mathrm{ref}}$, reference joint trajectories, $\bm q_{\mathrm{ip}}^{\mathrm{ref}}$, contact sequence $\sigma$, and optimal sampling rates for takeoff, flight, and landing phases, $dt^{t}$, $dt^{f}$, and $dt^{l}$. These trajectories are converted to 3D, $\bm{x}_c^{\mathrm{ref}},\: \bm q_{j}^{\mathrm{ref}}$, for real-time tracking control. 

Unsynchronized planning and control frequencies cause conflicts between the TO trajectory and the MPC prediction resolution. To address this, we pair the variable-model TO with a variable-frequency MPC that uses adaptive sampling frequencies and step lengths in its prediction horizon. This setup, combined with a joint-space PD controller, effectively tracks the jumping trajectories throughout the motion.
The resulting joint torques, $\bm \tau_j$, from controllers are sent to the robot motors. The feedback states from the robot includes CoM states, $\bm x_c = [\bm p_c; \bm \Theta; \bm \dot{\bm p}_c; \bm \omega]$ and joint states, $\bm q_j \in \mathbb R^{10}, \dot{\bm q}_j \in \mathbb R^{10}$. Where $\bm p_c \in \mathbb R^{3}$ is trunk CoM location, $\bm \Theta \in \mathbb R^{3}$ represents robot's Euler angles, $\bm \Theta = [\phi; \theta; \psi]$, $\dot{\bm p}_c \in \mathbb R^{3}$ is trunk CoM velocity, $\bm \omega \in \mathbb R^{3}$ is the world frame angular velocity.

\section{Proposed Approach}
\label{sec:approach}

In this section, we introduce the proposed approaches in this work, including the choice of variable dynamics models of the bipedal robot jumping problem, formulation of variable-model TO, and variable-frequency MPC.

\subsection{HECTOR Bipedal Robot }
\label{subsec:hector}

We use the HECTOR bipedal robot as our robot model and hardware experiment platform, introduced in the author's prior work \cite{li2023hector}.  HECTOR biped has 5-DoF legs with ankle actuation, stands at 70 $\unit{cm}$, and weighs 14 $\unit{kg}$. The biped can output 67.0 $\unit{Nm}$ of maximum torque at knee joints.
Detailed physical parameters are outlined in \cite{li2023hector}.


HECTOR's full joint-space dynamics equation of motion is described as follows. The joint-space generalized states ${\mathbf{q}} \in \mathbb{R}^{16}$ include $\bm p_c, \mathbf \Theta $, and $\bm q_j$.
\begin{align}
\label{eq:fullDynamics}
    \mathbf{H}(\mathbf{q})\ddot{\mathbf{q}} + \mathbf{C}(\mathbf{q}, \dot{\mathbf{q}}) = \mathbf{\Gamma} + \bm{J}_i(\mathbf{q})^\intercal \bm{\lambda}_i 
\end{align}

where $\mathbf{H} \in \mathbb{R}^{16 \times 16}$ is the mass-inertia matrix and $\mathbf{C} \in \mathbb{R}^{16}$ is the joint-space bias force term. $\mathbf{\Gamma} = [\mathbf{0}_6; \bm \tau_j]$ represents the actuation in the generalized coordinate. $\bm \lambda_i $ and $\bm{J}_i$ represent the external force applied to the system and its corresponding Jacobian matrix.

\subsection{Variable-fidelity Dynamics Modeling}
\label{subsec:dynamics}

In the proposed variable-model TO framework, we choose to use 3 different fidelities of dynamics models of bipedal robots for different phases of a jumping trajectory, illustrated in Fig. \ref{fig:controlArchi}. 
By leveraging variable models fidelities and trajectory resolutions, the TO can be efficiently solved online while providing hardware-realizable trajectories. 

We classify the bipedal jumping trajectory into three phases, as shown in Fig. \ref{fig:controlArchi}: (1) \textit{takeoff}, from the start of the jump until the feet leave the ground; (2) \textit{flight}, from when the robot is airborne until just before touchdown; and (3) \textit{landing}, from initial ground contact until the robot reaches the desired configuration.

\subsubsection{Takeoff Phase and 3-link Inverted Pendulum Model}
This phase is the most challenging part of the jumping, requiring strong lift-off while meeting actuation constraints. To ensure accuracy and optimize whole-body motion, we use a high-fidelity model with fine sampling steps. \cite{csomay2023nonlinear, nguyen2019optimized}.
We assume the biped remains in-contact with the ground before flight and model the planar takeoff motion using a full dynamics model of a fixed-base 3-link inverted pendulum in 2D. This fixed-base approach reduces computation while still applying virtual force constraints to represent the ground reaction forces of a floating-base model.
Each pendulum link has physical properties representing the corresponding biped link's mass and moment of inertia (MoI). The 2-D dynamics equation of motion (EoM) is as follows, 
\begin{align}
\label{eq:3linkDynamics}
      \mathbf{H}_{\mathrm{ip}}(\bm{q}_{\mathrm{ip}})\ddot{\bm{q}}_{\mathrm{ip}} + \mathbf{C}_{\mathrm{ip}}(\bm{q}_{\mathrm{ip}}, \dot{\bm{q}}_{\mathrm{ip}}) = \bm \tau_{\mathrm{ip}}
\end{align}
where $\bm{q}_{\mathrm{ip}} = [q_0, q_1, q_2]^\intercal$ describes the 3-link inverted pendulum joint states. $\bm \tau_{\mathrm{ip}}$ represents the 3-link model joint torques. 

\subsubsection{Flight Phase and MRBD}
In this phase, the TO must account for the effects of leg dynamics on whole-body angular momentum and adjust the position of the links for proper landing. However, since leg adjustments are minor in flight, we use a simplified multi-rigid-body model instead of full dynamics, treating each link's dynamics as external forces on the trunk CoM in 2D, 
$\bm p_c^{\mathrm{2D}}$ (\textit{e.g.}, three-particle model in \cite{yang2022balanced}). Note that we impose a virtual foot force $\bm F^v$ in dynamics formulation, to capture the effect of joint torques during the flight phase. The MRBD EoM is as follows,
\vspace{-0.1cm}
\begin{align}
\label{eq:MRBMDynamics}
      m_0 (\ddot{\bm p}_c^{\mathrm{2D}} + \bm g) = \bm F^{v} + \sum_{n=1}^{3} -m_n\bm g 
\end{align}
\vspace{-0.3cm}
\begin{align}
\label{eq:MRBMDynamics2}
      \frac{d}{dt}(I_0 \dot{\theta}) = (\bm p_f - \bm p_c^{\mathrm{2D}}) \times\bm F^{v} + \sum_{n=1}^{3} (\bm p_n - \bm p_c^{\mathrm{2D}}) \times -m_n \bm g 
\end{align}
where $m_n$ represents the mass of each link $n$, $n = 1,2,3$, $I_0$ represents the MoI of the trunk link,  $\theta$ represents the robot's pitch angle, $\bm p_n$ and $\bm p_f$ are the CoM location of each link and foot location in the world frame, accessed by forward kinematics (FK) of joint angles, $\texttt{FK}(\bm q_{\mathrm{ip}})$. 

\subsubsection{Landing Phase and SRBD}
As many legged-jumping TO frameworks have omitted, we choose to design the TO to include a landing and balancing phase of the jumping motion to generate a recovery trajectory for the MPC to track. Simplified float-based dynamics modeling is an effective approach in real-time control to allow soft impact \cite{wensing2017proprioceptive}. In addition, to establish a unified control framework and synchronize with the modeling of biped in our MPC, we design the landing part of the TO to consist of a single-rigid-body dynamics model, as described in detail in authors' prior work \cite{li2021force} and many other SRBD-based MPC on legged robots \cite{di2018dynamic, ding2022orientation}. The 2D EoM during landing is as follows, 
\begin{align}
\label{eq:SRBMDynamics}
      m_0 (\ddot{\bm p}_c^{\mathrm{2D}} + \bm g) = \bm F 
\end{align}
\vspace{-0.4cm}
\begin{align}
\label{eq:SRBMDynamics2}
      \frac{d}{dt}(I_c \dot{\theta}) = (\bm p_f - \bm p_c^{\mathrm{2D}}) \times \bm F + M 
\end{align}
where $\bm F$ is the 2D ground reaction force at the contact point, $M$ is the ground reaction moment in the x-y plane.

\subsection{Variable-model Trajectory Optimization}
\label{sec:amto}
 With the variable-fidelity models introduced, the following section will explain the details of the proposed variable-model TO and its NLP formulation. 

The Variable-model TO functions as an online trajectory planner for bipedal jumping and landing, using inputs such as jump distance, obstacle height, and landing location. It adapts model fidelity and trajectory resolution for each jumping phase and is formulated as a direct collocation TO problem solvable with an NLP solver.

 The optimization variables chosen in this TO are,
 \begin{align}
\label{eq:optvar}
\mathbf X = [\mathbf x_c, \:\bm q_{\mathrm{ip}}, \: \dot{\bm q}_{\mathrm{ip}}, \:\bm \tau_{\mathrm{ip}}, \:\bm F, \:M, \: dt^f, \:dt^l ]^\intercal;
\end{align}
where $dt^f$ is the flight phase step length and $dt^l$ is the landing phase step length. We choose to let the optimization determine the lengths of sampling steps during each phase, to prevent hard constraints on $dt$, which may lead to the optimization solving with infeasible contact timings or heavier computation.  Note that $\mathbf x_c = [p_{c,x}, p_{c,y}, \theta, \dot{p}_{c,x}, \dot{p}_{c,y}, \dot \theta]^\intercal$ is the 2D representation of the robot trunk CoM states $\bm x_c$. 

 The formulation of the finite horizon optimization problem with N steps is as follows,
\begin{alignat}{3}
\label{eq:cost}
\min_{\mathbf X} \quad & \sum_{i = 1}^{N} \Big\| \bm q_{\mathrm{ip},0} - \bm q_{\mathrm{ip},i}\Big\|^2 _{\bm Q_0} + \Big\| \dot{\bm q}_{\mathrm{ip},i}  \Big\|^2 _{\bm Q_1}  + \Big\| \bm \tau_{\mathrm{ip},i} \Big\|^2 _{\bm Q_2}\\ 
    \nonumber
    \textrm{s.t.} \quad & \quad
\end{alignat}
\vspace{-0.85cm}
\begin{subequations}
\setlength\abovedisplayskip{-3pt}
\begin{alignat}{3}
    \label{eq:qrange}
    \textrm{Joint angle:} \quad & \bm q_{\mathrm{min}} \leq \bm q_{\mathrm{ip}} \leq \bm q_{\mathrm{max}} \\
    \label{eq:dqrange}
    \textrm{Joint speed:} \quad & \dot{\bm q}_{\mathrm{min}} \leq \dot{\bm q}_{\mathrm{ip}} \leq \bm \dot{\bm q}_{\mathrm{max}} \\
    \label{eq:taurange}
     \textrm{Joint torque:} \quad &\bm \tau_{\mathrm{min}} \leq \bm \tau_{\mathrm{ip}} \leq \bm \tau_{\mathrm{max}} \\
     \label{eq:motorCons}
     \textrm{Motor Max Power:} \quad & P_{\mathrm{motor}} \leq 400 \unit{W}\\
    \label{eq:ic}
     \textrm{Initial Condition:} \quad & \mathbf{x}_{c,0} = \mathbf{x}_c^{\mathrm{IC}} \\
     \label{eq:tc}
    \: \textrm{Final Condition:} \quad & \mathbf{x}_{c,N} = \mathbf{x}_c^{\mathrm{FC}} \\ 
    \label{eq:FK}
    \textrm{Forward Kinematics:} \quad & \mathbf{x}_c = \texttt{FK}(\bm q_{\mathrm{ip}},\dot{\bm q}_{\mathrm{ip}})\\
    \label{eq:GRF}
    \textrm{Positive GRF:} \quad & 0 \leq F_y \\
    \label{eq:friction}
    \textrm{Friction:} \quad & -\mu  F_y \leq  F_x \leq \mu  F_y \\
    \label{eq:linefoot}
    \textrm{Line foot:} \quad & -l_h F_y \leq M \leq l_t F_y 
\end{alignat}
\end{subequations}
\begin{subequations}
\setlength\abovedisplayskip{-3pt}
\begin{alignat}{3}
    \hline 
    \nonumber
    \textbf{ Takeoff Phase} \quad & i = 0:N^{t}-1\\
    \label{eq:3linkDynamicsCons}
    \nonumber 
    \textrm{Dynamics (\ref{eq:3linkDynamics}):} \quad & \ddot{\bm{q}}_{\mathrm{ip},i} = \mathbf{H}_{\mathrm{ip},i}^{-1} (-\mathbf{C}_{\mathrm{ip},i} + \bm \tau_{\mathrm{ip}}) \\
    \nonumber 
    \: & \dot{\bm{q}}_{\mathrm{ip},i+1} = \dot{\bm{q}}_{\mathrm{ip},i} + \ddot{\bm{q}}_{\mathrm{ip},i}dt^t\\
    \: & \bm{q}_{\mathrm{ip},i+1} = \bm{q}_{\mathrm{ip},i} + \dot{\bm{q}}_{\mathrm{ip},i}dt^t \\
    \label{eq:support}
    \textrm{Support region:} \quad & -l_h \leq p_{\mathrm{rc},x}(\bm q_{\mathrm{ip}}) \leq l_t \\
    \label{eq:virtualGRF}
    \textrm{Virtual GRFM:} \quad & [\bm F_i; M_i]= {\bm J_c^\intercal}^{-1} \bm \tau_{\mathrm{ip},i}
\end{alignat}
\end{subequations}
\begin{subequations}
\setlength\abovedisplayskip{-3pt}
\begin{alignat}{3}
    \hline 
    \nonumber
    \textbf{ Flight Phase} \quad & i = N^{t}:N^{t}+N^{f}-1\\
    \label{eq:flightDynamicsCons}
    \textrm{Dynamics (\ref{eq:MRBMDynamics}-\ref{eq:MRBMDynamics2}):} \quad & \mathbf{x}_{c,i+1} = \mathbf{x}_{c,i}+\dot {\mathbf{x}}_{c,i}dt^f \\
    \label{eq:collisionsCons}
    \textrm{Collision avoidance:} \quad & h_{\mathrm{terrain},i} < p_{\mathrm{toe}, y}(\bm q_{\mathrm{ip},i})\\
    \label{eq:actuationCons}
    \textrm{Minimal actuation:} \quad & |\bm F^v_i| \leq 10\unit{N} \\
    \label{eq:flightdt}
    \textrm{Sampling rate:} \quad & 0 \leq dt^f
\end{alignat}
\end{subequations}
\begin{subequations}
\setlength\abovedisplayskip{-3pt}
\begin{alignat}{3}
    \hline 
    \nonumber
    \textbf{ Landing Phase} \quad & i = N^{t}+N^{f}: N-1\\
    \label{eq:landingDynamicsCons}
    \textrm{Dynamics (\ref{eq:SRBMDynamics}-\ref{eq:SRBMDynamics2}):} \quad & \mathbf{x}_{c,i+1} = \mathbf{x}_{c,i}+\dot {\mathbf{x}}_{c,i}dt^l \\
    \label{eq:landingdt}
    \textrm{Sampling rate:} \quad & 0.02\unit{s} \leq dt^l \leq 0.05\unit{s}
\end{alignat}
\end{subequations}

The objectives of the variable-model TO in (\ref{eq:cost}) contain minimizing joint movement, joint speed, and joint torque during the entire jumping motion. The goal is to generate a fluent jump with minimal actuation efforts. $\bm Q_0$, $\bm Q_1$, and $\bm Q_2$ are objective weighting matrices.  The NLP is subjected to several constraints. (\ref{eq:qrange}-\ref{eq:taurange}) describe the joint kinematics and actuation limits. (\ref{eq:motorCons}) describes the Unitree A1 motor power constraints based on its efficiency and wattage limit, which is an effective approach to ensure actuation awareness for hardware experiments \cite{nguyen2019optimized,chignoli2021humanoid}. (\ref{eq:ic}-\ref{eq:tc}) describes the initial and final conditions of the robot, based on the terrain setup and user commands. We use FK described in (\ref{eq:FK}) to bridge the joint space states $\bm q_{\mathrm{ip}}$ and robot trunk CoM states $\mathbf{x}_c$ and their discrete dynamics representations in variable models. (\ref{eq:GRF}-\ref{eq:friction}) ensure the ground reaction forces are always positive and respect friction constraints. (\ref{eq:linefoot}) describes the contact wrench cone constraint of line-foot \cite{caron2015stability}. 

Additional constraints are given and phase-specific, besides variable dynamics modeling. In (\ref{eq:virtualGRF}), we impose a virtual ground reaction forces $\bm F$ and moment $M$ from joint torques and virtual contact Jacobian to amplify the 3-link fix-base model to behave closely to a float-base model when in contact with the ground. (\ref{eq:support}) ensures the CoM of the entire robot, $\bm p_{\mathrm{rc}}$ stays within the support region, formed by the toe and heel length, $l_t, l_h$. During the flight phase, (\ref{eq:collisionsCons}) enforces the foot height of the robot $p_{\mathrm{toe},y}$ always above the terrain. (\ref{eq:actuationCons}) allows small virtual Cartesian foot forces (translated to joint torque actuations) to be applied to the dynamics, enabling the TO to adjust the robot's feet for better landing configuration.

\subsection{Variable-frequency MPC}

Once the variable-model TO generates an optimal jumping trajectory online, the subsequential variable-frequency MPC will work in conjunction with joint PD control to track the reference trajectories $\bm x_c^{\mathrm{ref}}$ and $\bm q_j^{\mathrm{ref}}$, which are mapped directly from optimization solution $\mathbf X$. 
The MPC uses the planned contact sequence, variable trajectory frequency, and sampling step length $dt$ to synchronize sampling frequencies with the variable-model TO to fully leverage trajectory resolutions.
\begin{figure*}[!t]
\vspace{0.2cm}
     \centering
        \includegraphics[clip, trim=0cm 9.8cm 0cm 0cm, width=2\columnwidth]{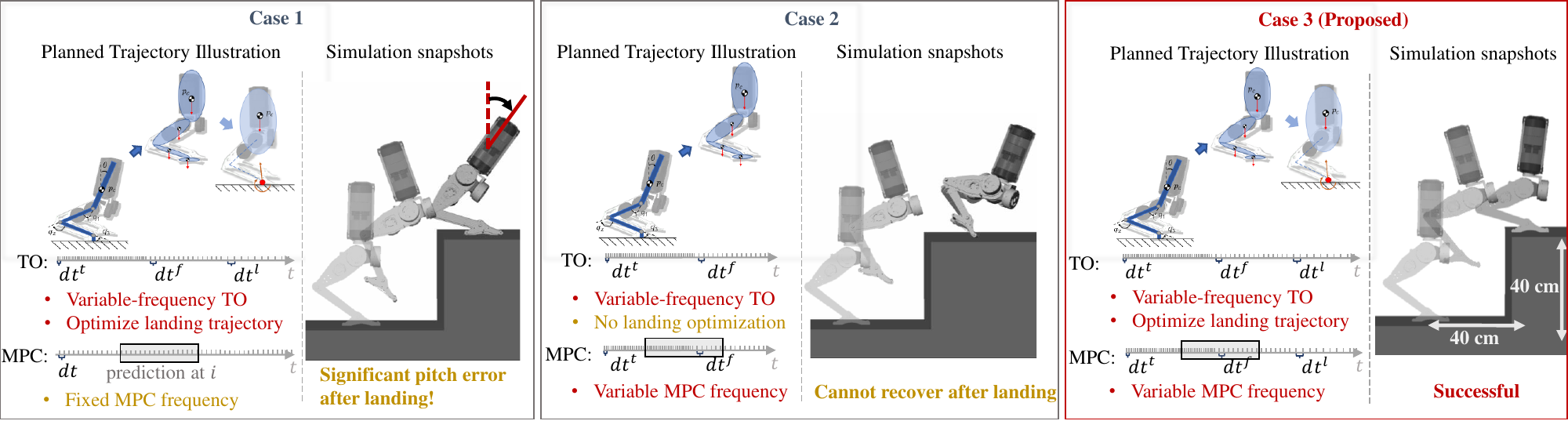}
     \caption{{\bfseries{Comparing Baseline Approaches with Proposed Approach in Simulation. }} Red text represents proposed methods in this paper while yellow text represents baseline methods. } 
    \label{fig:compare}
    \vspace{-0.2cm}
\end{figure*}

\begin{figure}[!t]
\vspace{0.0cm}
     \centering
        \includegraphics[clip, trim=4.2cm 9.5cm 4.2cm 9.5cm, width=1\columnwidth]{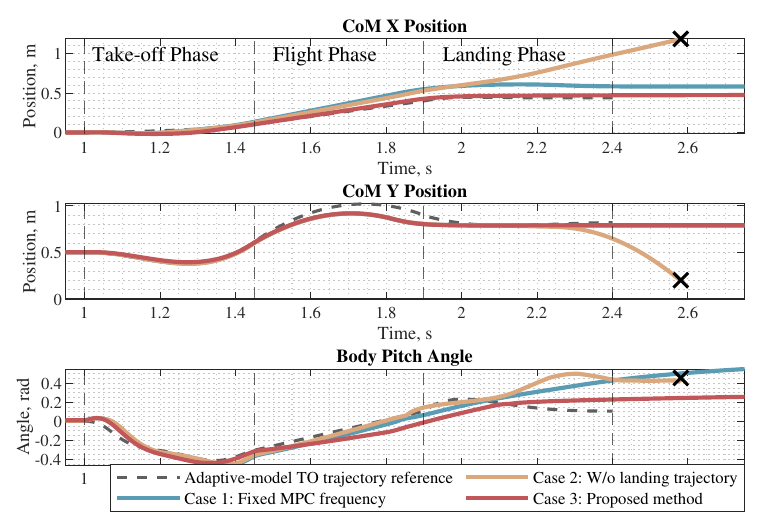}
 \caption{{\bfseries{CoM Trajectory Tracking Plots. }} Associated with comparative analysis shown in Fig.~\ref{fig:compare}} 
    \label{fig:sim_tracking}
    \vspace{-0.2cm}
\end{figure}

The variable-frequency MPC leverages the SRBD modeling of bipedal robot HECTOR, with the assumption of negligible effect of leg movement during landing control. The dynamics formulation closely follows the author's prior work \cite{li2023dynamic}, a 3D extension of (\ref{eq:SRBMDynamics}-\ref{eq:SRBMDynamics2}). We choose to include gravity $\bm g$ in the optimization variable $\bm x = [\bm x_c; \: \bm g]$ to linearize the dynamics and form the discrete-time state-space equation at time step $k$, $\bm {x}_{k+1} = \bm {\hat{A}}_{d,k}\bm x_k + \bm {\hat{B}}_{d,k}\bm u_k$,
where ${\hat{\bm A_d} \in \mathbb{R}^{15\times15}}$ and ${\hat{\bm B_d} \in \mathbb{R}^{15\times12}}$ are discrete-time state-space dynamics matrices. 
 The control input $\bm u$ includes the 3D ground reaction forces and moments of both foot, $\bm u = [\bm F_0;\bm F_1;\bm M_0;\bm M_1]$ The formulation of the optimization problem with finite horizon $h$ is as follows, 
\begin{alignat}{3}
\label{eq:MPCcost}
\min_{\bm x, \bm u} \quad & \sum_{i = 0}^{h-1} \Big\| \bm x_k-  \bm x^{\mathrm{ref}}_k\Big\|^2 _{\bm R_0} + \Big\| \bm{u}_k  \Big\|^2 _{\bm R_1}\\ 
    \nonumber
    \textrm{s.t.} \quad & \quad
\end{alignat}
\vspace{-0.35cm}
\begin{subequations}
\allowdisplaybreaks
\setlength\abovedisplayskip{-3pt}
\begin{alignat}{3}
    \label{eq:mpcDynamics}
    \textrm{Dynamics: } \quad & \bm {x}_{k+1} = \bm {\hat{A}}_{d,k}\bm x_k + \bm {\hat{B}}_{d,k}\bm u_k  \\
    \label{eq:mpcFriction}
    \textrm{Friction pyramid} \quad & -\mu  {F}_{z,n} \leq  F_{x,n} \leq \mu  {F}_{z,n} \\
    \nonumber
    \textrm{of foot $n$:} \quad & -\mu {F}_{z,n} \leq  F_{y,n} \leq \mu  {F}_{z,n} \\
    \label{eq:MPCforce}
    \textrm{Force limit:} \quad & 0 \leq  F_{z,n} \leq {F}_{\mathrm{max}} \\
    \label{eq:Mx}
    \textrm{Moment X:} \quad & \prescript{}{\mathcal{B}}{M_{x,n}} = 0 \\
    \label{eq:MPClinefoot}
    \textrm{Line foot:} \quad & -l_h F_{z,n} \leq M_{y,n} \leq l_t F_{z,n}
\end{alignat}
\end{subequations}
The objective of the problem is to drive state $\bm x$ close to the TO reference and minimize control input $\bm u$. These objectives are weighted by diagonal matrices $\bm R_0\in  \mathbb{R}^{15\times15}$ and $\bm R_1\in \mathbb{R}^{12\times12}$. (\ref{eq:Mx}) enforces foot moment in the x-direction is always zeros in the robot's body frame, as discussed in \cite{li2021force}. 

\subsection{Real-time Jumping Control}

The real-time jumping control consists of both variable-frequency MPC for tracking CoM trajectory and joint-PD control for tracking the joint trajectory simultaneously. The joint torque command is computed as,
\begin{align}
\label{eq:jumpingControl}
\bm \tau_j = \bm K_p (\bm q_j^{\mathrm{ref}}-\bm q_j) + \bm K_d (\dot{\bm q}_j^{\mathrm{ref}}-\dot{\bm q}_j) + \bm J_c^\intercal \bm u
\end{align}
where $\bm J_c$ is the contact points' world frame Jacobian matrix.

To achieve dynamic continuous jumping, we integrate the variable-model TO and variable-frequency MPC by using variable step lengths $dt$ to create preview trajectories for each phase. The MPC connects each jump's end configuration to the next jump's start, guiding the robot to prepare for the following jump \cite{nguyen2022continuous}. Unlike single jumps, continuous jumping leverages conserved energy by taking off from a lowered height, with the variable-frequency MPC adjusting loop frequency and sampling step lengths accordingly.

\section{Results}
\label{sec:Results}

This section presents the hardware results of the proposed Variable-model Optimization framework in bipedal jumping, including comparisons with baseline approaches and robust behaviors in single and continuous jumps. 

\subsection{Implementing Details}
 The variable-model TO is set up as NLP problem and is solved with CasADi toolbox \cite{Andersson2019} and \texttt{FATROP} solver \cite{vanroye2023fatrop}. In simulation validation, we use the HECTOR open-source simulation software in Simulink \cite{hectorGithub}. On the hardware, the real-time MPC is solved as a quadratic programming (QP) problem via \texttt{qpOASES} solver in C++. The detailed hardware control parameters are shown in Table. \ref{tab:parameters}.

\begin{table}[!h]
\centering
\begin{scriptsize}
\vspace{-0.2cm}
\centering
\caption{Experiment Parameters}
\label{tab:parameters}
  \begin{tabular}{ | m{12em} | m{12em} |  } 
  \hline
  \hline
  $\bm Q_0$ = 0.1 & $N^t$ = 70, \: $dt^t$ = 0.01s\\ 
  \hline
  $\bm Q_1$ = 0.5 & $N^f$ = 30 \\ 
  \hline
  $\bm Q_2$ = 0.01 & $N^l$ = 30\\ 
  \hline
  $\mu$ = 0.5& $h$ = 10\\ 
  \hline
  \hline
  \multicolumn{2}{| c |}{$\bm R_0 = \texttt{diag}([300\: 300\: 400\: 150\: 200\: 150\: 1\: 1\: 1\: 1\: 1\: 10 \:0])$} \\
  \multicolumn{2}{| c |}{$\bm R_1 = \texttt{diag}([1\: 1\: 1\: 1\: 1\: 1\: 1\: 1\: 1\: 1\: 1 \: 1 ]\cdot 10^{-6})$ \quad \quad \quad \quad \quad} \\
  \hline
  \multicolumn{2}{| c |}{$\bm K_p = \texttt{diag}([150\: 150\: 150\: 150\: 100])$} \\
  \multicolumn{2}{| c |}{$\bm K_d = \texttt{diag}([2\: 2\: 2\: 2\: 2\: 1 ])$ \quad \quad \quad \quad \quad} \\
  \hline
  \hline
\end{tabular}
\end{scriptsize}
\end{table}
\vspace{-0.2cm}

\subsection{Comparison of Solve Time}
We set up bipedal jumping TO problems with several different dynamics model fidelities, including the full dynamics model, kinodynamics model, SRBD model, and the proposed variable-model scheme. The solve time comparisons are presented in Table.\ref{tab:comparison}, under 3 different tasks setups: (1) 25 \unit{cm} forward leap, (2) 40 \unit{cm} forward leap, and (3) 40 \unit{cm} forward and 40 \unit{cm} high jump (Fig. \ref{fig:compare}).
With tailored model choices during different jumping phases and variable sampling frequencies, our method can achieve real-time computation of TO (\textit{i.e.}, when TO solve time is shorter than the jumping motion length, TO can generate the next-jump trajectory before the end of the current jump). Even though SRBD-based TO can be solved very efficiently through QP solver, the trajectory accuracy still poses challenges to being employed in dynamic jumping motions \cite{nguyen2022contact}.

\begin{table}[!t]
\vspace{0.2cm}
\centering
\setlength{\tabcolsep}{3.5pt}
\caption{Bipedal Jumping TO Solve Time Comparison}
\label{tab:comparison}
\begin{scriptsize}
  \begin{tabular}{ | m{6em} | m{6em} | m{6em} | m{6em} | m{6em} | } 
  \hline
  \hline
  \makecell[c]{Model} & \makecell[c]{Full-order} & \makecell[c]{Kinodynamics} & \makecell[c]{SRBD} & \makecell[c]{\textbf{Proposed}}\\ 
  \hline
\hline
  Dimension & \makecell[c]{2-D} & \makecell[c]{3-D} & \makecell[c]{3-D} &  \makecell[c]{\textbf{2-D}} \\ 
  \hline
  Ref. work$^{a}$ & \makecell[c]{\cite{posa2014direct,nguyen2019optimized}} & \makecell[c]{\cite{chignoli2021humanoid,zhang2023design}} & \makecell[c]{\cite{nguyen2022contact,li2023dynamic}} &  \makecell[c]{\textbf{-}} \\ 
  \hline
  $\#$ of steps & \makecell[c]{100} & \makecell[c]{100} & \makecell[c]{100} &  \makecell[c]{\textbf{100+30$^{b}$}} \\ 
  \hline
  Step length & \makecell[c]{0.01 \unit{s}} & \makecell[c]{0.01 \unit{s}} & \makecell[c]{0.01 \unit{s}} &  \makecell[c]{\textbf{variable}}\\ 
  \hline
  Solver & \makecell[c]{\texttt{FATROP}} & \makecell[c]{\texttt{FATROP}} & \makecell[c]{\texttt{qpOASES}} & \makecell[c]{\texttt{\textbf{FATROP}}} \\ 
  \hline
  \hline
  Case & \multicolumn{4}{ c |}{Solve time} \\
  \hline
  25 \unit{cm} F& \makecell[c]{6.9 \unit{s}} & \makecell[c]{4.5 \unit{s}} & \makecell[c]{0.09 \unit{s}} &  \makecell[c]{\textbf{0.91 \unit{s}}}\\ 
  \hline
  40 \unit{cm} F& \makecell[c]{7.5 \unit{s}} & \makecell[c]{3.9 \unit{s}} & \makecell[c]{0.09 \unit{s}} &  \makecell[c]{\textbf{0.84 \unit{s}}}\\ 
  \hline
  40 \unit{cm} F$\&$H & \makecell[c]{7.2 \unit{s}} & \makecell[c]{4.3 \unit{s}} & \makecell[c]{0.11 \unit{s}} &  \makecell[c]{\textbf{0.93 \unit{s}}}\\ 
  \hline
  \hline
\end{tabular}
\end{scriptsize}
\begin{flushleft}
\footnotesize{\scriptsize{$^a$ The compared methods based on these references are re-constructed to fit our robot model and problem setup. \\  $^b$ Refers to the additional landing trajectory in proposed method. }}    
\end{flushleft}
\vspace{-0.3cm}
\end{table}

\begin{figure}[!t]
\vspace{0.2cm}
     \centering
        \includegraphics[clip, trim=4cm 10.5cm 4cm 10.5cm, width=1\columnwidth]{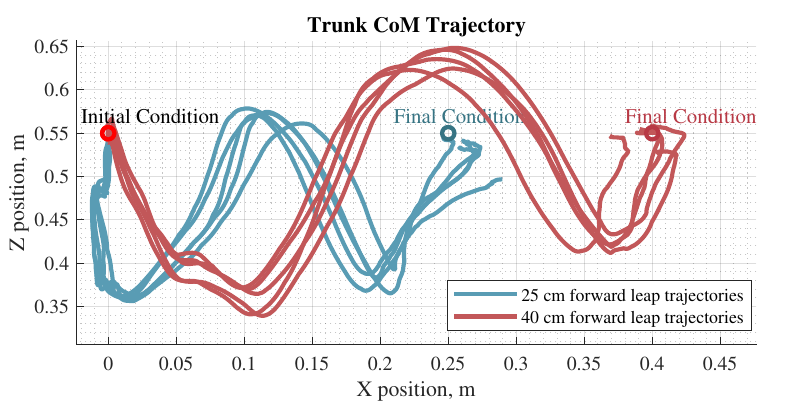}
 \caption{{\bfseries{Jumping Trajectory Overlay. }} CoM trajectories overlaid in jumping repeatability tests.} 
    \label{fig:overlay}
    \vspace{-0.2cm}
\end{figure}

\subsection{Comparison of Baseline Approaches}
We demonstrate the significance of landing trajectory and variable-frequency frameworks on bipedal jumping. In Fig. \ref{fig:compare}, we compare baseline approaches with our method in simulations. Case 1 shows that using a fixed-frequency MPC with a variable-frequency trajectory (\textit{i.e.}, mismatched resolutions) leads to significant pitch error post-impact. Case 2 demonstrates that omitting the landing recovery trajectory optimization leads to instability after impact. Our approach in Case 3 successfully enables the robot to jump and land on a 40 cm platform.

\subsection{Jumping Repeatability}
To validate the repeatability of the jumping behaviors with the proposed framework, we run 53 jumping experiments, which include (1) 18 trials of 40-cm forward leaps and (2) 35 trials of 25-cm forward leaps. Five of each jumping trajectories are overlaid in Fig. \ref{fig:overlay} to showcase the consistency of performance with our control framework. 

\subsection{Landing Robustness}
\begin{figure}[!t]
\vspace{0.2cm}
     \centering
        \includegraphics[clip, trim=0cm 8cm 6.3cm 0cm, width=1\columnwidth]{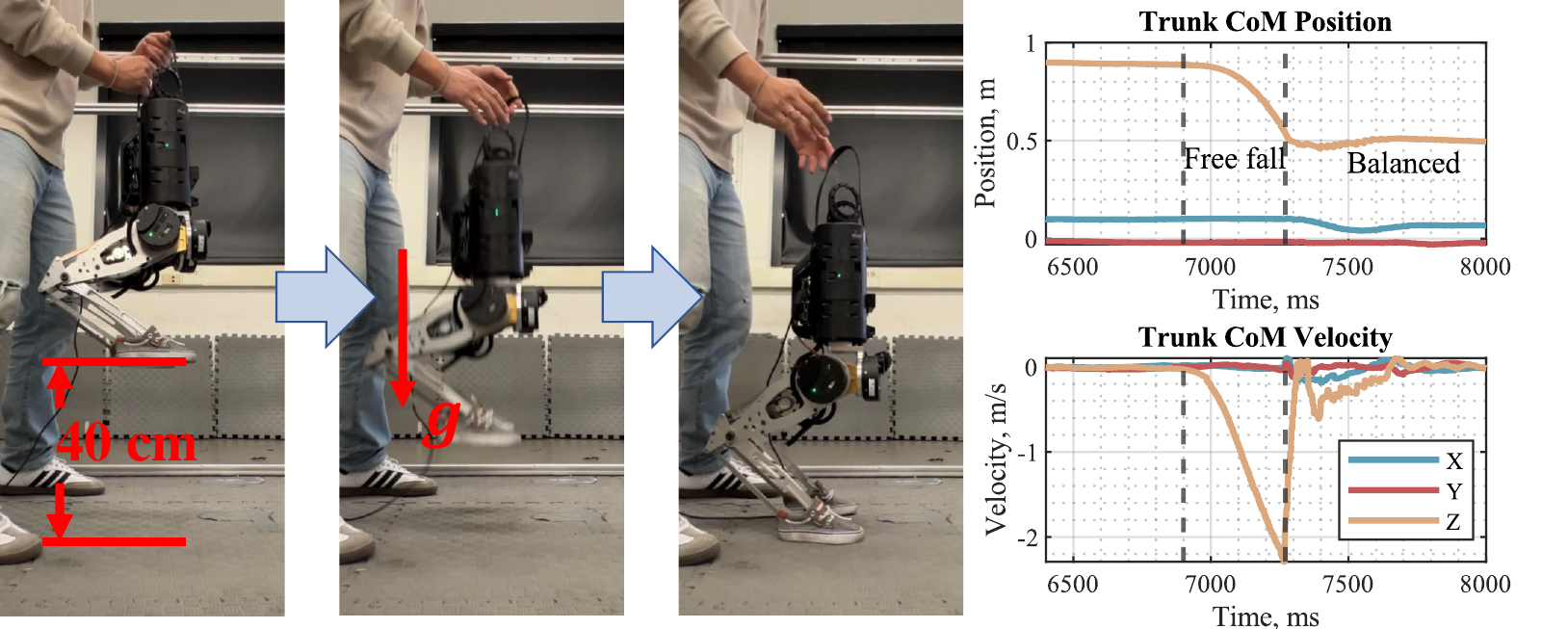}
      \caption{{\bfseries{Balancing after 0.4 m Freefall. }} Experimental snapshots and CoM plots.}
    \label{fig:freefall}
    \vspace{-0.2cm}
\end{figure}

A successful jumping control cannot omit the importance of its landing recovery. In this work, we leverage the SRBD-based variable-frequency MPC for impact absorption and recovery after landing. We verify the robustness of the landing control by manually dropping the robot from a height of 40 cm foot clearance off the ground (57$\%$ of robot height), shown in Fig. \ref{fig:freefall}. The robot can effectively absorb the impact and high z-direction velocity after freefall and balance immediately. 

\subsection{Robustness over Challenging Terrain}
We successfully applied our control method to dynamic continuous jumps, handling terrain perturbations up to 20 cm (Fig. \ref{fig:jumpdown}) and 5 cm height changes between jumps (Fig.~\ref{fig:terrain}). The robot, without prior knowledge of disturbances, navigates these terrains effectively. Additionally, the robot can jump over wooden boxes with increasing heights, up to 20 cm, in under 6 seconds. Joint tracking and torque performance are shown in Fig. \ref{fig:title}, demonstrating effective tracking within actuation limits.

\begin{figure}[!t]
     \centering
        \includegraphics[clip, trim=4cm 10.5cm 4cm 10.5cm, width=1\columnwidth]{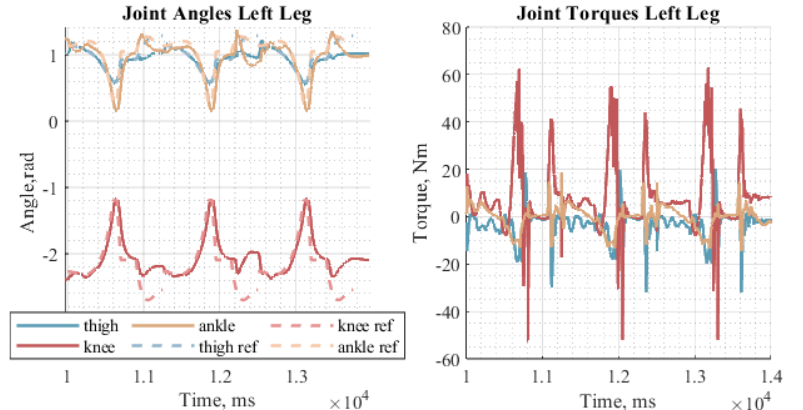}
     \caption{{\bfseries{Continous Jumping over Discrete Terrain Experiment Joint Plots. }} Left leg joint position tracking and joint torque plots. Note that the hip yaw and hip roll joints are omitted for comparison in this plot due to sagittal motion.} 
    \label{fig:box}
    \vspace{-0.2cm}
\end{figure}

\section{Conclusion and Future Work}
\label{sec:Conclusion}

In conclusion, we present a variable-model optimization framework for real-time planning and control of bipedal jumping. We demonstrated the effectiveness of variable model selection and trajectory optimization, real-time application feasibility, the need for landing recovery trajectories, and the importance of matching frequencies and sampling steps between the planned trajectory and tracking MPC, through extensive hardware experimentation.
The ongoing refinement of the framework involves: (1) adapting the 2-D inverted pendulum model to a 3-D framework, potentially using a double multi-link model with closed-loop kinematics for efficient 3-D motion; and (2) integrating perception for \textit{one-step preview} terrain data to enhance autonomous planning in hardware experiments.

\newpage
\balance
\bibliographystyle{ieeetr}
\bibliography{reference.bib}

\end{document}